\documentclass[runningheads]{llncs}
\usepackage{amsmath}
\usepackage{amssymb}
\usepackage{graphicx}
\usepackage{graphicx}
\usepackage{hyperref}
\begin{document}
\title{Curating Subject ID Labels using Keypoint Signatures}
%
%
\author{Laurent Chauvin \inst{1} \and
Matthew Toews \inst{1}}
%
%
\institute{\textsuperscript{1}\'Ecole de Technologie Sup\'erieure, Montreal, Canada\\ \email{laurent.chauvin0@gmail.com}}
%
\maketitle              

\begin{figure}[htbp]
 \centering
   \includegraphics[width=.86\linewidth]{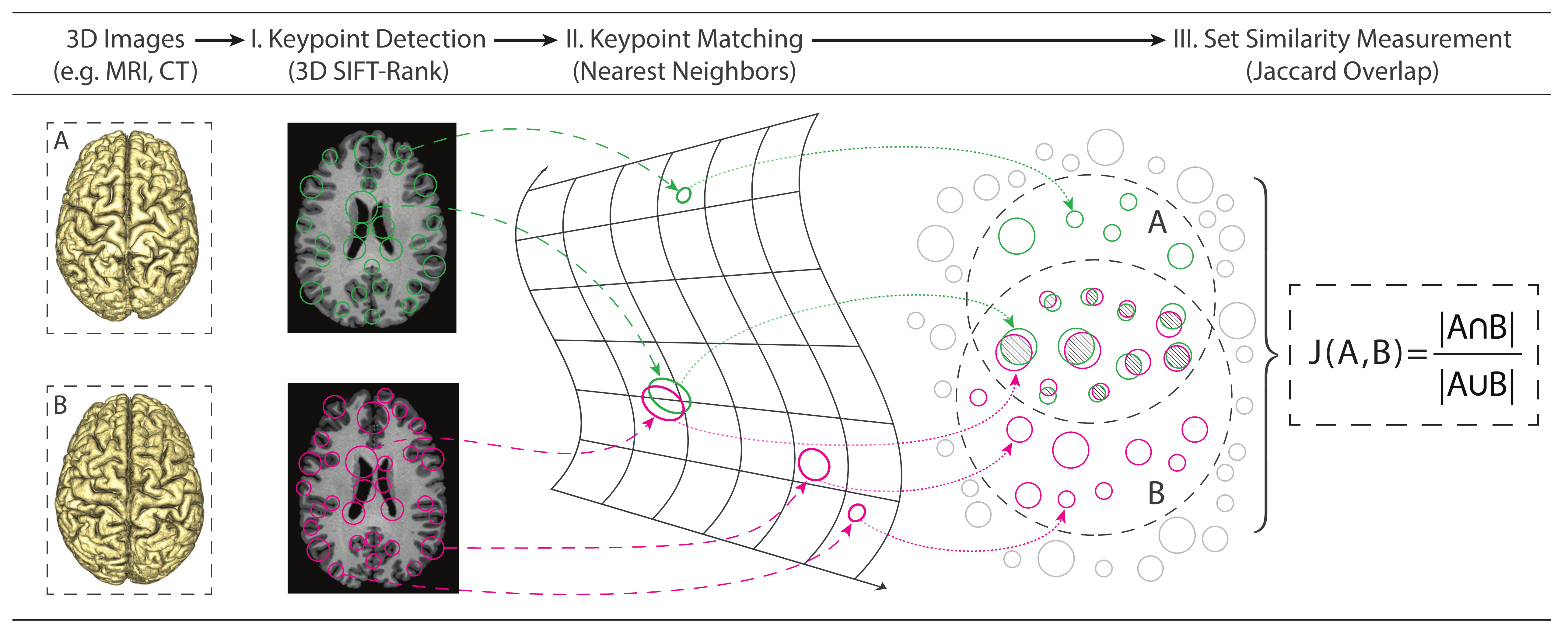}
   \caption{Computing the log Jaccard distance $d(A,B) = -\log J(A,B)$ between a pair of generic medical image volumes $(A,B)$ involves (left to right) SIFT-Rank keypoint detection~\cite{Toews2013a}, K-NN keypoint matching and Jaccard overlap evaluation.}
   \label{fig:workflow}
\end{figure}

\begin{figure}[htbp]
 \centering
    \includegraphics[width=1\linewidth]{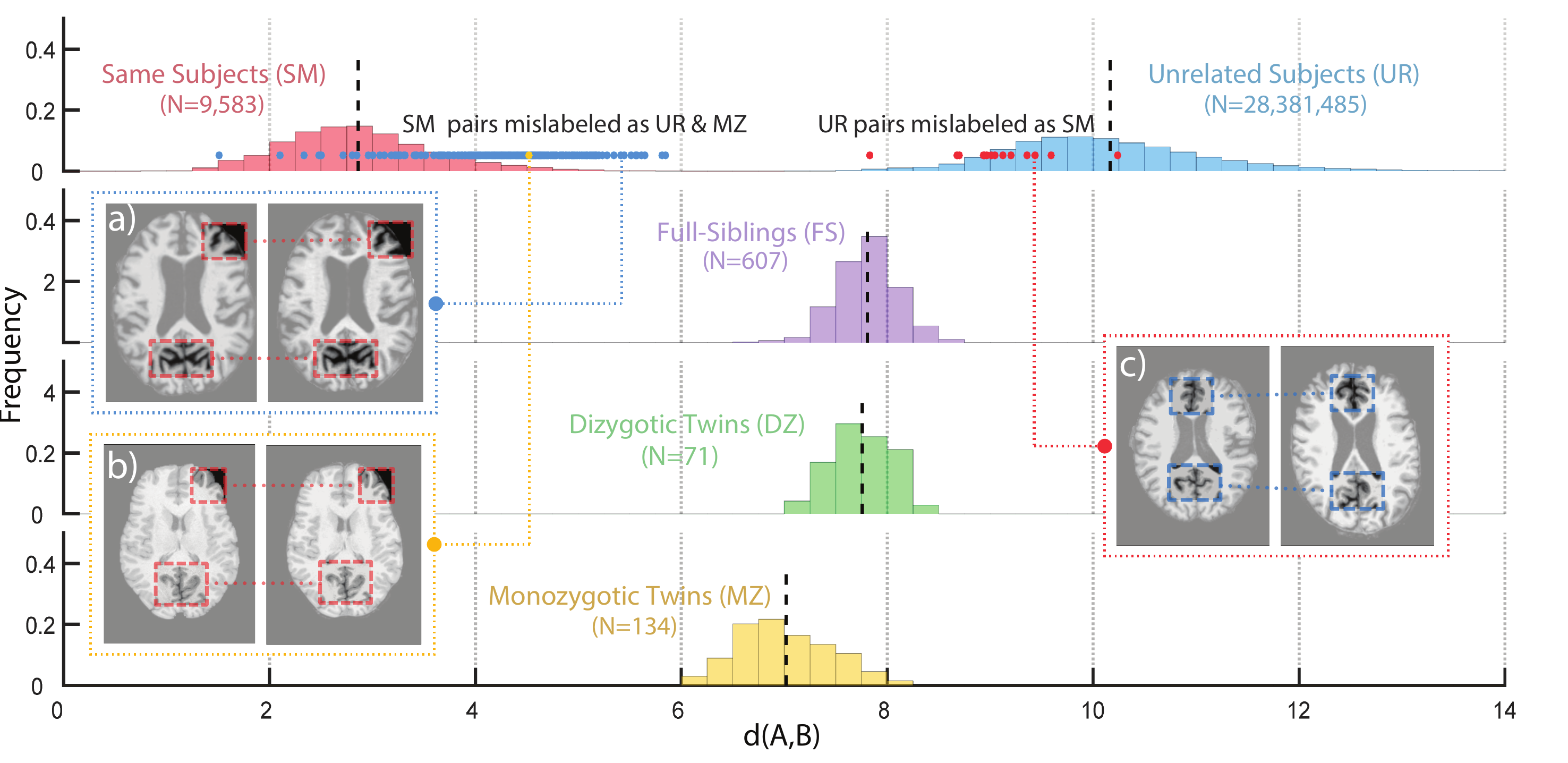}
    \caption{Distributions of the pairwise log Jaccard distance $d(A,B)$ conditional on relationship labels, dots indicate obvious label inconsistencies, with visual examples of same subject images mislabelled as different (a,b) and different subject images labeled as the same (c).}
    \label{fig:results}
\end{figure}

Subject ID labels are unique, anonymized codes that can be used to group all images of a subject while maintaining anonymity. ID errors may be inadvertently introduced manually error during enrollment and may lead to systematic error into machine learning evaluation (e.g. due to double-dipping) or potential patient misdiagnosis in clinical contexts. 
Here we describe a highly efficient system for curating subject ID labels in large generic medical image datasets, based on the 3D image keypoint representation, which recently led to the discovery of previously unknown labeling errors in widely-used public brain MRI datasets~\cite{Chauvin2020NeuroimageRelatives}.\\
\noindent \\
{\bf \underline{Method}:} ID errors are identified using a pair-wise (dis)similarity measure $d(A,B)$ to evaluate and flag all image pairs $(A,B)$ that are unexpectedly similar or dissimilar. The primary computational challenge is the $O(N^2)$ algorithmic complexity of pairwise comparisons for a set of $N$ images, which becomes intractable as $N \rightarrow \infty$. $d(A,B)$ is defined by the log of the Jaccard distance~\cite{Levandowsky1971} between keypoint sets $A=\{f_i\}$ and $B=\{f_j\}$, as illustrated in Figure~\ref{fig:workflow}. Salient 3D SIFT-Rank keypoints~\cite{Toews2013a} are detected in individual images, where each keypoint is a generic local anatomical pattern and is represented as a compact 64-element descriptor derived from rank-ordered gradient histogram elements~\cite{Toews2009}.  Enumerating nearest neighbor (NN) descriptor matches, i.e. equivalent anatomical patterns between $N^2$ image pairs is achieved in $O(N~log~N)$ complexity via approximate K-NN search using the KD-tree structure~\cite{Muja2014}.
\\
\noindent
{\bf \underline{Results}:}
A total of $N=7536$ MRIs of $3334$ unique subjects are pooled from four large public neuroimaging databases and used in evaluation: the Human Connectome Project (HCP)~\cite{VanEssen2012}, the Open Access Series of Imaging Studies (OASIS) 1 and 3~\cite{Marcus2007}, and the Alzheimer's Disease Neuroimaging Initiative (ADNI)~\cite{Jack2008}. Each of $N(N-1)/2=28,391,880$ image pairs $(A,B)$, is associated with a unique pair-wise relationship label: same subjects (SM), monozygotic twins (MZ), dizygotic twins (DZ), full-sibling (FS), and unrelated (UR), images in different databases are assumed to be unrelated. As expected, pairwise distance distributions in Figure~\ref{fig:results} are lowest and highest for the same (SM, pink) and unrelated subjects (UR, blue), respectively, with twin and non-twin sibling distances falling in between. Obvious outliers (blue, pink points) reveal labeling errors (see Figure~\ref{fig:results}), which were visually confirmed to be either images of the same subject mislabeled as different (blue dots) or images of different subjects labeled as the same (pink).

\noindent
{\bf \underline{Discussion}:} For a typical $256^3$-voxel image, keypoint detection requires 1-2 seconds per image via a GPU implementation~\footnote{\url{https://github.com/3dsift-rank/3DSIFT-Rank}} and produces 1000-5000 keypoints per image depending on the content. Enumerating all NN images requires 100-350 milliseconds per image. The memory footprint for keypoints is 100x smaller than image data and could be incorporated as a DICOM image header. The HCP has since removed duplicate subjects.

%
%
%
\bibliographystyle{splncs04}
\bibliography{ref}
\end{document}